\documentclass[runningheads]{llncs}
\usepackage{graphicx}
\usepackage{listings}
\usepackage{perpage}
\usepackage{courier}
\usepackage{makecell}
\usepackage[ampersand]{easylist}
\usepackage{amssymb}
\usepackage{multirow}
\ListProperties(Hide=100, Hang=true, Progressive=3ex, Style*=-- ,
Style2*=$\bullet$ ,Style3*=$\circ$ ,Style4*=\tiny$\blacksquare$ )
\MakePerPage{footnote}
\begin{document}
\pagenumbering{arabic}
\pagestyle{plain}
\title{Recognizing Musical Entities in User-generated Content}
%
%
\author{Lorenzo Porcaro\inst{1} \and
Horacio Saggion\inst{2}}
\authorrunning{Lorenzo Porcaro \and Horacio Saggion}

%
\institute{Music Technology Group, Universitat Pompeu Fabra   \and
TALN Natural Language Processing Group, Universitat Pompeu Fabra}
\maketitle              
\begin{abstract}
Recognizing Musical Entities is important for Music Information Retrieval (MIR) since it can improve the performance of several tasks such as music recommendation, genre classification or artist similarity. However, most entity recognition systems in the music domain have concentrated on formal texts (e.g. artists’ biographies, encyclopedic articles, etc.), ignoring rich and noisy user-generated content. In this work, we present a novel method to recognize musical entities in Twitter content generated by users following a classical music radio channel. Our approach takes advantage of both formal radio schedule and users’ tweets to improve entity recognition. We instantiate several machine learning algorithms to perform entity recognition combining task-specific and corpus-based features. We also show how to improve recognition results by jointly considering formal and user-generated content. 

\keywords{Named Entity Recognition  \and Music Information Retrieval \and User-generated Content.}
\end{abstract}
\section{Introduction}
The increasing use of social media and microblogging services has broken new ground in the field of Information Extraction (IE) from user-generated content (UGC). Understanding the information contained in users' content has become one of the main goal for many applications, due to the uniqueness and the variety of this data~\cite{ref_proc1}. However, the highly informal and noisy status of these sources makes it difficult to apply techniques proposed by the NLP community for dealing with formal and structured content~\cite{ref_proc2}. 

In this work, we analyze a set of tweets related to a specific classical music radio channel, \textit{BBC Radio 3}\footnote{www.twitter.com/BBCRadio3}, interested in detecting two types of musical named entities, \textit{Contributor} and \textit{Musical Work}.

The method proposed makes use of the information extracted from the radio schedule for creating links between users' tweets and tracks broadcasted. Thanks to this linking, we aim to detect when users refer to entities included into the schedule. Apart from that, we consider a series of linguistic features, partly taken from the NLP literature and partly specifically designed for this task, for building statistical models able to recognize the musical entities. To that aim, we perform several experiments with a supervised learning model, Support Vector Machine (SVM), and a recurrent neural network architecture, a bidirectional LSTM with a CRF layer (biLSTM-CRF).

The  contributions in this work are summarized as follows:
\begin{itemize}
\item  A method to recognize musical entities from user-generated content which combines contextual information (i.e. radio schedule) with Machine Learning models for improving the accuracy while recognizing the entities.
\item The release of  language resources such as an user-generated and  bot-generated Twitter corpora manually annotated, usable for both MIR and NLP researches, and domain specific word embeddings.
\end{itemize}
The paper is structured as follows. In Section 2, we present a review of the previous works related to Named Entity Recognition, focusing on its application on UGC and MIR. Afterwards, in Section 3 it is presented the methodology of this work, describing the dataset and the method proposed. In Section 4, the results obtained are shown. Finally, in Section 5 conclusions are discussed.

\section{Related Work}
Named Entity Recognition (NER), or alternatively Named Entity Recognition and Classification (NERC), is the task of detecting entities in an input text and to assign them to a specific class. It starts to be defined in the early '80, and over the years several approaches have been proposed~\cite{ref_article1}. Early systems were based on handcrafted rule-based algorithms, while recently several contributions by Machine Learning scientists have helped in integrating probabilistic models into NER systems. 

In particular, new developments in neural architectures have become an important resource for this task. Their main advantages are that they do not need language-specific knowledge resources~\cite{ref_proc3}, and they are robust to the noisy and short nature of social media messages~\cite{ref_proc4}. Indeed, according to a performance analysis of several Named Entity Recognition and Linking systems presented in~\cite{ref_article2}, it has been found that poor capitalization is one of the main issues when dealing with microblog content. Apart from that, typographic errors and the ubiquitous occurrence of out-of-vocabulary (OOV) words also cause drops in NER recall and precision, together with shortenings and slang, particularly pronounced in tweets.

\begin{table}
\centering
\caption{Examples of user-generated tweets.}\label{tab1}
\begin{tabular}{|l|l|}
\hline
1 &No Schoenberg or Webern?? Beethoven is there but not his pno sonata op. 101??\\
\hline
2 &Heard some of Opera 'Oberon' today... Weber... Only a little....\\
\hline
3 &Cavalleria Rusticana...hm..from a Competition that very nearly didn't get entered!\\
\hline
\end{tabular}
\end{table}

Music Information Retrieval (MIR) is an interdisciplinary field which borrows tools of several disciplines, such as signal processing, musicology, machine learning, psychology and many others, for extracting knowledge from musical objects (be them audio, texts, etc.)~\cite{ref_book1}. In the last decade, several MIR tasks have benefited from NLP, such as sound and music recommendation~\cite{ref_article3}, automatic summary of song review~\cite{ref_proc5}, artist similarity~\cite{ref_proc6} and genre classification~\cite{ref_proc7}. 

In the field of IE, a first approach for detecting musical named entities from raw text, based on Hidden Markov Models, has been proposed in~\cite{ref_proc8}. In~\cite{ref_proc9}, the authors combine state-of-the-art Entity Linking (EL) systems to tackle the problem of detecting musical entities from raw texts. The method proposed relies on the \textit{argumentum ad populum} intuition, so if two or more different EL systems perform the same prediction in linking a named entity mention, the more likely this prediction is to be correct. In detail, the off-the-shelf systems used are: DBpedia Spotlight~\cite{ref_proc10}, TagMe~\cite{ref_article4}, Babelfy~\cite{ref_article5}. Moreover, a first Musical Entity Linking, MEL\footnote{http://mel.mtg.upf.edu}  has been presented in~\cite{ref_proc11} which combines different state-of-the-art NLP libraries and SimpleBrainz, an RDF knowledge base created from MusicBrainz\footnote{https://musicbrainz.org} after a simplification process.

Furthermore, \textit{Twitter} has also been at the center of many studies done by the MIR community. As example, for building a music recommender system~\cite{ref_proc12} analyzes tweets containing keywords like \textit{nowplaying} or \textit{listeningto}. In~\cite{ref_proc6}, a similar dataset it is used for discovering cultural listening patterns. Publicly available \textit{Twitter} corpora built for MIR investigations have been created, among others the \textit{Million Musical Tweets} dataset\footnote{http://www.cp.jku.at/datasets/MMTD}~\cite{ref_proc13} and the \textit{\#nowplaying} dataset\footnote{http://dbis-nowplaying.uibk.ac.at}~\cite{ref_proc14}.

\section{Methodology}
We propose a hybrid method which recognizes musical entities in UGC using both contextual and linguistic information. We focus on detecting two types of entities: 
\begin{easylist}
& \textit{Contributor}: person who is related to a musical work (composer, performer, conductor, etc).
& \textit{Musical Work}: musical composition or recording (symphony, concerto, overture, etc).
\end{easylist}
\begin{table}
\centering
\caption{Example of entities annotated and corresponding formal forms, from the user-generated tweet (1) in Table 1.}\label{tab10}
\begin{tabular}{|l|l|}
\hline
\multicolumn{1}{|c|}{\makecell{\textit{Informal form}}} & \multicolumn{1}{c|}{\makecell{\textit{Formal form}}}  \\
\hline
\makecell{Schoenberg} & \makecell{Arnold Franz Walter Schoenberg} \\

\makecell{Webern} & \makecell{Anton Friedrich Wilhelm Webern} \\

\makecell{Beethoven} & \makecell{Ludwig Van Beethoven} \\

\makecell{\hspace{15mm}pno sonata op. 101 \hspace{15mm}\hspace{15mm}} &\makecell{Piano Sonata No. 28 in A major, Op. 101}\\
\hline
\end{tabular}
\end{table}

As case study, we have chosen to analyze tweets extracted from the channel of a classical music radio, \textit{BBC Radio 3}. The choice to focus on classical music has been mostly motivated by the particular discrepancy between the informal language used in the social platform and the formal nomenclature of contributors and musical works. Indeed, users when referring to a musician or to a classical piece in a tweet, rarely use the full name of the person or of the work, as shown in Table 2.

We extract information from the radio schedule for recreating the musical context to analyze user-generated tweets, detecting when they are referring to a specific work or contributor recently played. We manage to associate to every track broadcasted a list of entities, thanks to the tweets automatically posted by the \textit{BBC Radio3 Music Bot}\footnote{https://twitter.com/BBCR3MusicBot}, where it is described the track actually on air in the radio. In Table 3, examples of bot-generated tweets are shown.

\begin{table}
\centering
\caption{Examples of bot-generated tweets.}\label{tab2}
\begin{tabular}{|l|l|}
\hline
1 & \makecell{Now Playing Joaqu\'in Rodrigo, Goran Listes - 3 Piezas espa\~nolas for guitar \\ \#joaqu\'inrodrigo,\#goranlistes} \\
\hline
2 &  \makecell{Now Playing Robert Schumann, Luka Mitev - Phantasiestücke, \\Op 73  \#robertschumann,\#lukamitev}\\
\hline
3 & \makecell{Now Playing Pyotr Ilyich Tchaikovsky, MusicAeterna - Symphony No.6 in B \\ minor \#pyotrilyichtchaikovsky, \#musicaeterna} \\
\hline
\end{tabular}
\end{table}

Afterwards, we detect the entities on the user-generated content by means of two methods: on one side, we use the entities extracted from the radio schedule for generating candidates entities in the user-generated tweets, thanks to a matching algorithm based on time proximity and string similarity. On the other side, we create a statistical model capable of detecting entities directly from the UGC, aimed to model the informal language of the raw texts. In Figure 1, an overview of the system proposed is presented.

\begin{figure}
\centering
\includegraphics[width=\textwidth]{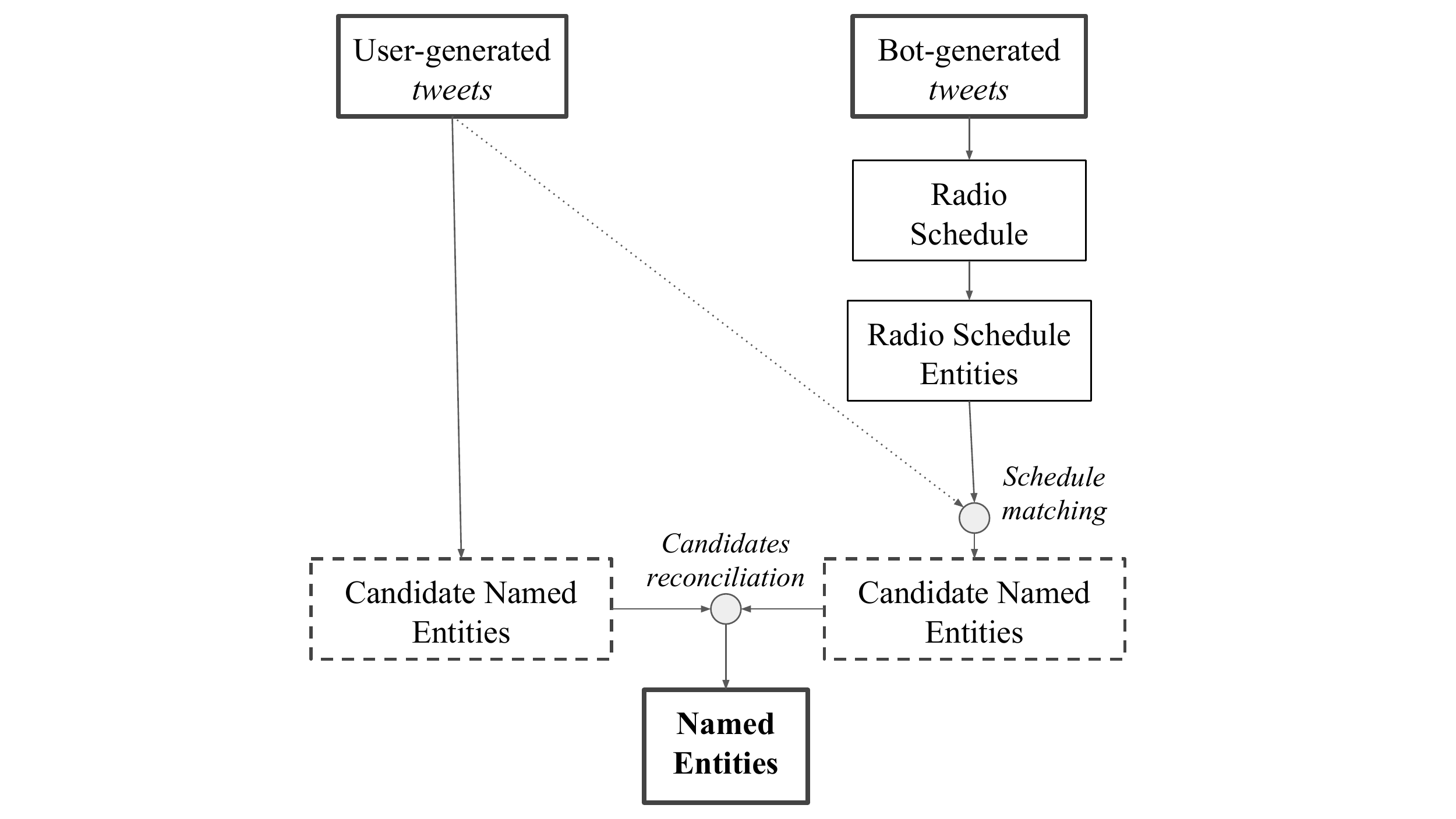}
\caption{Overview of the NER system proposed} \label{fig1}
\end{figure}

\subsection{Dataset}
In May 2018, we crawled Twitter using the Python library \textit{Tweepy}\footnote{http://www.tweepy.org}, creating two datasets on which \textit{Contributor} and \textit{Musical Work} entities have been manually annotated, using IOB tags.

The first set contains user-generated tweets related to the \textit{BBC Radio 3} channel. It represents the source of user-generated content on which we aim to predict the named entities. We create it filtering the messages containing hashtags related to \textit{BBC Radio 3}, such as \textit{\#BBCRadio3} or \textit{\#BBCR3}. We obtain a set of 2,225 unique user-generated tweets. The second set consists of the messages automatically generated by the \textit{BBC Radio 3 Music Bot}. This set contains 5,093 automatically generated tweets, thanks to which we have recreated the schedule\footnote{https://github.com/LPorcaro/musicner}. 

In Table 4, the amount of tokens and relative entities annotated are reported for the two datasets. For evaluation purposes, both sets are split in a training part (80\%) and two test sets (10\% each one) randomly chosen. Within the user-generated corpora, entities annotated are only about 5\% of the whole amount of tokens. In the case of the automatically generated tweets, the percentage is significantly greater and entities represent about the 50\%. 
\begin{table}
\centering
\caption{Tokens' distributions within the two datasets: user-generated tweets (top) and bot-generated tweets (bottom)}\label{tab9}
\begin{tabular}{|l|r|r|r|}
\hline
&\multicolumn{1}{c|}{Training}&\multicolumn{1}{c|}{TestA}&\multicolumn{1}{c|}{TestB}\\
\hline
\multicolumn{4}{|l|}{User-generated tweets}\\
\hline
\textit{Contributor}&1.069 (3,12\%)&	119 (2,96\%)&	127 (2,97\%)\\
\textit{Musical Work}&964 (2,81\%)&	118 (2,93\%)&	163 (3,81\%)\\
\hline
Total tokens&34.247	&4.016&	4.275\\
\hline
\hline
\multicolumn{4}{|l|}{Bot-generated tweets}\\
\hline
\textit{Contributor}&15.162 (27,50\%)&	1.852 (22,93\%)	&1.879 (27,30\%)\\
\textit{Musical Work}&12.904 (23,40\%)&	1.625 (23,56\%)	&1.689 (24,48\%)\\
\hline
Total tokens&55.122	&6.897&	6.881\\
\hline
\end{tabular}
\end{table}

\subsection{NER system}
According to the literature reviewed, state-of-the-art NER systems proposed by the NLP community are not tailored to detect musical entities in user-generated content. Consequently, our first objective has been to understand how to adapt existing systems for achieving significant results in this task. 

\begin{table}
\centering
\caption{Example of musical named entities annotated}\label{tab3}
\begin{tabular}{|l|l|l|l|l|l|l|l|l|l|}
\multicolumn{1}{c}{\makecell{Beethoven}} &\multicolumn{1}{c}{\makecell{is}} &\multicolumn{1}{c}{there} &\multicolumn{1}{c}{but} &\multicolumn{1}{c}{not}& \multicolumn{1}{c}{his}& \multicolumn{1}{c}{\makecell{pno}} &\multicolumn{1}{c}{\makecell{sonata}}& \multicolumn{1}{c}{\makecell{op.}} &\multicolumn{1}{c}{\makecell{101}}\\
\hline
\multicolumn{1}{c}{B-CONTR} &\multicolumn{1}{c}{O}& \multicolumn{1}{c}{\makecell{O}}&\multicolumn{1}{c}{\makecell{O}} &\multicolumn{1}{c}{\makecell{O}}& \multicolumn{1}{c}{\makecell{O}} &\multicolumn{1}{c}{B-WORK} &\multicolumn{1}{c}{I-WORK} &\multicolumn{1}{c}{I-WORK} &\multicolumn{1}{c}{I-WORK}\\
\end{tabular}
\end{table}

In the following sections, we describe separately the features, the word embeddings and the models considered. All the resources used are publicy available\footnote{https://github.com/LPorcaro/musicner}.

\subsubsection{Features' description} 
We define a set of features for characterizing the text at the token level. We mix standard linguistic features, such as Part-Of-Speech (POS) and chunk tag, together with several gazetteers specifically built for classical music, and a series of features representing tokens' left and right context. For extracting the POS and the chunk tag we use the Python library \textit{twitter\_nlp}\footnote{https://github.com/aritter/twitter\_nlp}, presented in~\cite{ref_proc2}. 

In total, we define 26 features for describing each token: 1)POS tag; 2)Chunk tag; 3)Position of the token within the text, normalized between 0 and 1; 4)If the token starts with a capital letter; 5)If the token is a digit. Gazetteers: 6)Contributor first names;  7)Contributor last names; 8)Contributor types ("soprano", "violinist", etc.);  9)Classical work types ("symphony", "overture", etc.); 10)Musical instruments; 11)Opus forms ("op", "opus"); 12)Work number forms ("no", "number"); 13)Work keys ("C", "D", "E", "F" , "G" , "A", "B", "flat", "sharp"); 14)Work Modes ("major", "minor", "m"). Finally, we complete the tokens' description including as token's features the surface form, the POS and the chunk tag of the previous and the following two tokens (12 features).

\subsubsection{Word embedding}
We consider two sets of GloVe word embeddings~\cite{ref_proc15} for training the neural architecture, one pre-trained with 2B of tweets, publicy downloadable\footnote{https://github.com/stanfordnlp/GloVe}, one trained with a corpora of 300K tweets collected during the 2014-2017 \textit{BBC Proms Festivals} and disjoint from the data used in our experiments.

\subsubsection{Models}
The first model considered for this task has been the John Platt's sequential minimal optimization algorithm for training a support vector classifier~\cite{ref_article6}, implemented in WEKA~\cite{ref_article7}. Indeed, in~\cite{ref_article8} results shown that SVM outperforms other machine learning models, such as Decision Trees and Naive Bayes, obtaining the best accuracy when detecting named entities from the user-generated tweets.

However, recent advances in Deep Learning techniques have shown that the NER task can benefit from the use of neural architectures, such as biLSTM-networks~\cite{ref_proc3,ref_proc4}. We use the implementation\footnote{https://github.com/UKPLab/emnlp2017-bilstm-cnn-crf} proposed in~\cite{ref_proc16} for conducting three different experiments. In the first, we train the model using only the word embeddings as feature. In the second, together with the word embeddings we use the POS and chunk tag. In the third, all the features previously defined are included, in addition to the word embeddings. For every experiment, we use both the pre-trained embeddings and the ones that we created with our \textit{Twitter} corpora. In section 4, results obtained from the several experiments are reported.

\subsection{Schedule matching}
The bot-generated tweets present a predefined structure and a formal language, which facilitates the entities detection. In this dataset, our goal is to assign to each track played on the radio, represented by a tweet, a list of entities extracted from the tweet raw text. For achieving that, we experiment with the algorithms and features presented previously, obtaining an high level of accuracy, as presented in section 4. The hypothesis considered is that when a radio listener posts a tweet, it is possible that she is referring to a track which has been played a relatively short time before. In this cases, we want to show that knowing the radio schedule can help improving the results when detecting entities.

Once assigned a list of entities to each track, we perform two types of matching. Firstly, within the tracks we identify the ones which have been played in a fixed range of time (\textit{t}) before and after the generation of the user's tweet. Using the resulting tracks, we create a list of candidates entities on which performing string similarity. The score of the matching based on string similarity is computed as the ratio of the number of tokens in common between an entity and the input tweet, and the total number of token of the entity:

\begin{equation}
score_{string\hspace{1mm}matching}(Entity) = \frac{\#(TokensEntity\cap input\hspace{1mm} tweet)}{\#TokensEntity}
\end{equation}

In order to exclude trivial matches, tokens within a list of stop words are not considered while performing string matching. The final score is a weighted combination of the string matching score and the time proximity of the track, aimed to enhance matches from tracks played closer to the time when the user is posting the tweet. 

The performance of the algorithm depends, apart from the time proximity threshold \textit{t}, also on other two thresholds related to the string matching, one for the \textit{Musical Work} (\textit{w}) and one for the \textit{Contributor} (\textit{c}) entities. It has been necessary for avoiding to include candidate entities matched against the schedule with a low score, often source of false positives or negatives. Consequently, as last step \textit{Contributor} and \textit{Musical Work} candidates entities with respectively a string matching score lower than \textit{c} and \textit{w}, are filtered out.  In Figure 2, an example of \textit{Musical Work} entity recognized in an user-generated tweet using the schedule information is presented.

\begin{figure}
\includegraphics[width=\textwidth]{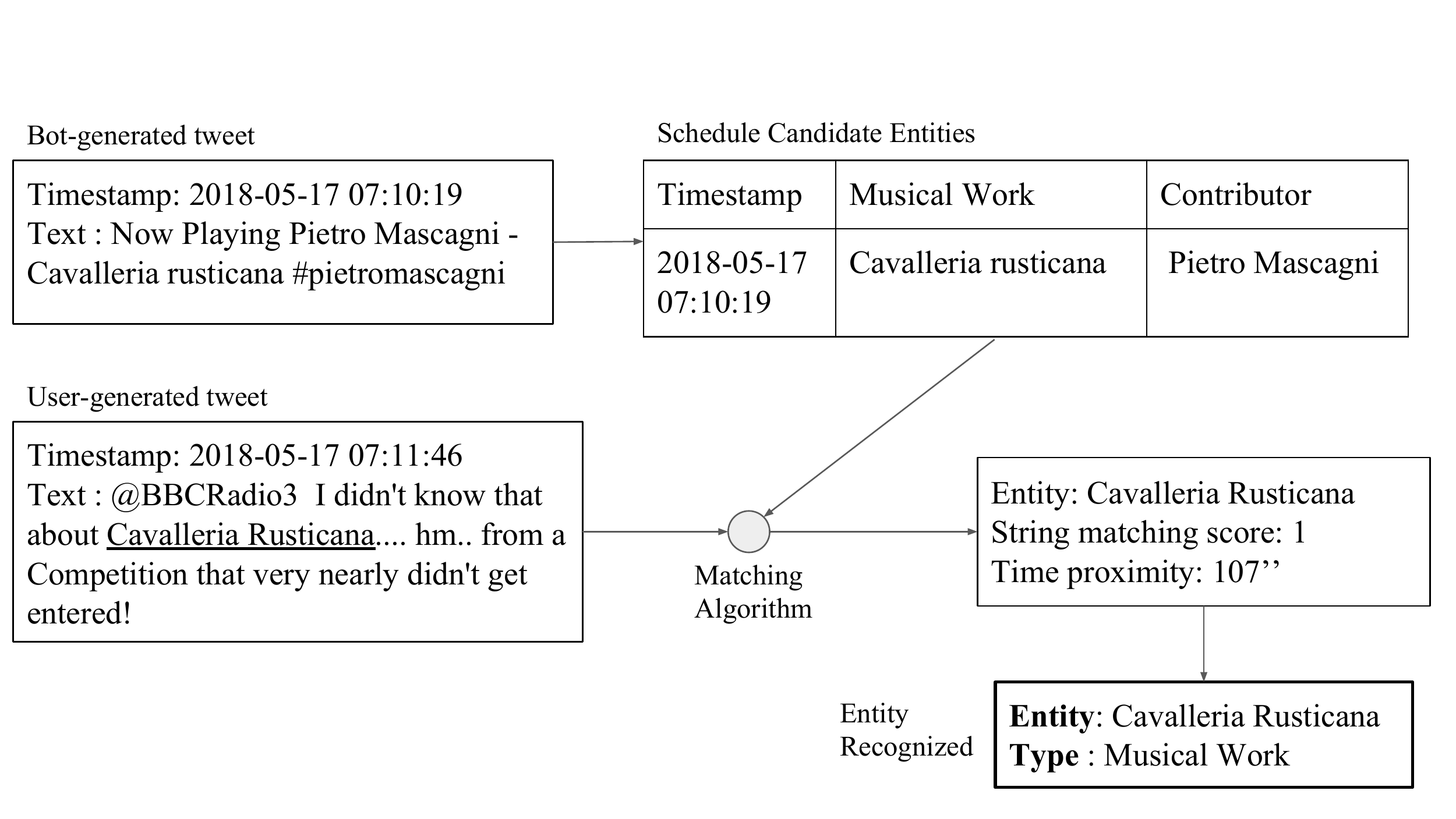}
\caption{Example of the workflow for recognizing entities in UGC using the information from the radio schedule} \label{fig3}
\end{figure}

\subsubsection{Candidates Reconciliation}
The entities recognized from the schedule matching are joined with the ones obtained directly from the statistical models. In the joined results, the criteria is to give priority to the entities recognized from the machine learning techniques. If they do not return any entities, the entities predicted by the schedule matching are considered. Our strategy is justified by the poorer results obtained by the NER based only on the schedule matching, compared to the other models used in the experiments, to be presented in the next section.

\section{Results}
The performances of the NER experiments are reported separately for three different parts of the system proposed. 

\begin{table}
\centering
\caption{F1 score for \textit{Contributor}(C) and \textit{Musical Work}(MW) entities recognized from bot-generated tweets (top) and user-generated tweets (bottom)}\label{tab6}
\begin{tabular}{|l|l|l|l|l|l|l|l|l|}
\hline
\makecell{Model}&\makecell{Features}&\makecell{GloVe vectors}&\multicolumn{2}{l}{\makecell{Training}}&\multicolumn{2}{|l|}{\makecell{TestA}}&\multicolumn{2}{|l|}{\makecell{TestB}}\\
\hline
\multicolumn{3}{|l|}{\textit{Bot-generated tweets}}&\multicolumn{1}{c}{\textit{C}}&\makecell{\textit{MW}}&\multicolumn{1}{c}{\textit{C}}&\makecell{\textit{MW}}&\multicolumn{1}{c}{\textit{C}}&\makecell{\textit{MW}}\\
\hline
SVM & all & -- & \multicolumn{1}{c}{99.12}&\multicolumn{1}{c|}{97.70}&\multicolumn{1}{c}{97.74}&\multicolumn{1}{c|}{94.32}&\multicolumn{1}{c}{97.88}&\multicolumn{1}{c|}{95.42}\\

biLSTM-CRF & -- & trained & \multicolumn{1}{c}{98.95}&\multicolumn{1}{c|}{97.07}&\multicolumn{1}{c}{98.06}&\multicolumn{1}{c|}{92.99}&\multicolumn{1}{c}{98.33}&\multicolumn{1}{c|}{95.59}\\

 & & pre-trained & \multicolumn{1}{c}{99.34}&\multicolumn{1}{c|}{94.94}&\multicolumn{1}{c}{97.88}&\multicolumn{1}{c|}{91.40}&\multicolumn{1}{c}{98.27}&\multicolumn{1}{c|}{92.35}\\

biLSTM-CRF & POS+chunk & trained & \multicolumn{1}{c}{\textbf{99.94}}&\multicolumn{1}{c|}{98.28}&\multicolumn{1}{c}{97.99}&\multicolumn{1}{c|}{\textbf{94.68}}&\multicolumn{1}{c}{98.03}&\multicolumn{1}{c|}{\textbf{95.97}}\\

 &  & pre-trained & \multicolumn{1}{c}{99.69}&\multicolumn{1}{c|}{97.23}&\multicolumn{1}{c}{98.12}&\multicolumn{1}{c|}{93.30}&\multicolumn{1}{c}{98.49}&\multicolumn{1}{c|}{93.61}\\

biLSTM-CRF & all & trained &\multicolumn{1}{c}{99.80} &\multicolumn{1}{c|}{98.22}&\multicolumn{1}{c}{97.70}&\multicolumn{1}{c|}{91.99}&\multicolumn{1}{c}{98.36}&\multicolumn{1}{c|}{94.48}\\

 &  & pre-trained &\multicolumn{1}{c}{99.90} &\multicolumn{1}{c|}{\textbf{99.40}}&\multicolumn{1}{c}{\textbf{98.24}}&\multicolumn{1}{c|}{90.46}&\multicolumn{1}{c}{\textbf{98.78}}&\multicolumn{1}{c|}{94.23}\\
\hline
\hline
\multicolumn{3}{|l}{\textit{User-generated tweets}}&\multicolumn{6}{|l|}{}\\
\hline
SVM & all & -- &\multicolumn{1}{c}{95.44}  &\multicolumn{1}{c|}{80.80}& \multicolumn{1}{c}{64.91}&\multicolumn{1}{c|}{\textbf{33.48}}& \multicolumn{1}{c}{61.02}&\multicolumn{1}{c|}{36.21}\\

biLSTM-CRF & -- & trained &\multicolumn{1}{c}{79.09}  &\multicolumn{1}{c|}{51.51}&\multicolumn{1}{c}{60.00}&\multicolumn{1}{c|}{26.66}&\multicolumn{1}{c}{67.02}&\multicolumn{1}{c|}{31.48}\\

 & & pre-trained &\multicolumn{1}{c}{85.51}  &\multicolumn{1}{c|}{69.28} &\multicolumn{1}{c}{70.00} &\multicolumn{1}{c|}{33.33} &\multicolumn{1}{c}{\textbf{71.26}} &\multicolumn{1}{c|}{32.08} \\

biLSTM-CRF & POS+chunk & trained & \multicolumn{1}{c}{79.37}&\multicolumn{1}{c|}{50.90}&\multicolumn{1}{c}{61.23}&\multicolumn{1}{c|}{28.98}&\multicolumn{1}{c}{62.03}&\multicolumn{1}{c|}{\textbf{40.00}}\\

 &  & pre-trained & \multicolumn{1}{c}{73.51}&\multicolumn{1}{c|}{37.28}&\multicolumn{1}{c}{\textbf{71.62}}&\multicolumn{1}{c|}{25.00}&\multicolumn{1}{c}{63.74}&\multicolumn{1}{c|}{25.53}\\

biLSTM-CRF & all & trained & \multicolumn{1}{c}{97.42} & \multicolumn{1}{c|}{\textbf{88.92}}& \multicolumn{1}{c}{66.22}& \multicolumn{1}{c|}{28.17}& \multicolumn{1}{c}{69.11}& \multicolumn{1}{c|}{36.36}\\

 &  & pre-trained & \multicolumn{1}{c}{\textbf{98.46}}&\multicolumn{1}{c|}{87.35}&\multicolumn{1}{c}{68.79}&\multicolumn{1}{c|}{23.68}&\multicolumn{1}{c}{70.41}&\multicolumn{1}{c|}{29.51}\\
\hline
\end{tabular}

\end{table}

Table 6 presents the comparison of the various methods while performing NER on the bot-generated corpora and the user-generated corpora. Results shown that, in the first case, in the training set the F1 score is always greater than 97\%, with a maximum of 99.65\%. With both test sets performances decrease, varying between 94-97\%. In the case of UGC, comparing the F1 score we can observe how performances significantly decrease. It can be considered a natural consequence of the complex nature of the users' informal language in comparison to the structured message created by the bot.

In Table 7, results of the schedule matching are reported. We can observe how the quality of the linking performed by the algorithm is correlated to the choice of the three thresholds. Indeed, the \textit{Precision} score increase when the time threshold decrease, admitting less candidates as entities during the matching, and when the string similarity thresholds increase, accepting only candidates with an higher degree of similarity. The behaviour of the \textit{Recall} score is inverted.

\begin{table}
\centering
\caption{\textit{Precision} (P), \textit{Recall} (R) and \textit{F1} score for \textit{Contributor} (C) and \textit{Musical Work} (MW) of the schedule matching algorithm. \textit{w} indicates the \textit{Musical Work} string similarity threshold, \textit{c} indicates the \textit{Contributor} string similarity threshold and \textit{t} indicates the time proximity threshold in seconds}\label{tab5}
\begin{tabular}{|l|l|l|l|l|l|l|l|l|l|l|}
\cline{1-11}
\multicolumn{2}{|c|}{}&\multicolumn{3}{l}{\makecell{\textit{t}=800}}&\multicolumn{3}{|l|}{\makecell{\textit{t}=1000}}&\multicolumn{3}{l|}{\makecell{\textit{t}=1200}}\\
\cline{3-11}
\multicolumn{2}{|c|}{}&\multicolumn{1}{c}{P} & \multicolumn{1}{c}{R} &\multicolumn{1}{l|}{\makecell{F1}}&\multicolumn{1}{c}{P} & \multicolumn{1}{c}{R} &\multicolumn{1}{l|}{\makecell{F1}}&\multicolumn{1}{c}{P} & \multicolumn{1}{c}{R} &\multicolumn{1}{l|}{\makecell{F1}} \\
\hline
w=0.33, c=0.33&\textit{C} &  \multicolumn{1}{c}{72.49}&\multicolumn{1}{c}{16.49}&\makecell{26.87}&	\multicolumn{1}{c}{69.86}&\multicolumn{1}{c}{17.57}	&28.08&	\multicolumn{1}{c}{68.66}&	\multicolumn{1}{c}{\textbf{17.93}}&\textbf{28.43}\\

&\textit{MW}&\multicolumn{1}{c}{26.42}	&\multicolumn{1}{c}{4.78}&\makecell{8.10}&\multicolumn{1}{c}{26.05}	&	\multicolumn{1}{c}{5.29}&\makecell{8.79}&\multicolumn{1}{c}{23.66}&\multicolumn{1}{c}{5.29}	&\makecell{8.65}\\
\hline
w=0.33, c=0.5&\textit{C} & \multicolumn{1}{c}{\textbf{76.77}}&\multicolumn{1}{c}{14.32}&24.14&	\multicolumn{1}{c}{74.10}&	\multicolumn{1}{c}{15.64}&25.83&\multicolumn{1}{c}{73.89}&	\multicolumn{1}{c}{16.00}&\makecell{26.30}\\

 & \textit{MW}&\multicolumn{1}{c}{27.1}&\multicolumn{1}{c}{4.95}&\makecell{8.37}&\multicolumn{1}{c}{26.67}	&	\multicolumn{1}{c}{5.46}&\textbf{9.06}&	\multicolumn{1}{c}{24.24}&\multicolumn{1}{c}{\textbf{5.46}}& \makecell{8.91}	\\
\hline
w=0.5, c=0.5 & \textit{C}& \multicolumn{1}{c}{\textbf{76.77}}&\multicolumn{1}{c}{14.32}	&24.14&\multicolumn{1}{c}{74.71}	&\multicolumn{1}{c}{15.64}&25.87&\multicolumn{1}{c}{73.89}&	\multicolumn{1}{c}{16.00}&26.30\\

 &\textit{MW}&\multicolumn{1}{c}{ \textbf{30.43}}&\multicolumn{1}{c}{4.78}&\makecell{8.26}&	\multicolumn{1}{c}{30.30}&\multicolumn{1}{c}{5.12}&\makecell{8.76}&	\multicolumn{1}{c}{27.52}&	\multicolumn{1}{c}{5.12}&\makecell{8.63}\\
\hline
\end{tabular}
\end{table}

\begin{table}
\centering
\caption{\textit{Precision} (P), \textit{Recall} (R) and \textit{F1} score for \textit{Contributor} (C) and \textit{Musical Work} (MW) entities recognized from user-generated tweets using the biLSTM-CRF network together with the schedule matching. The thresholds used for the matching are t=1200, w=0.5, c=0.5.}\label{tab8}. 
\begin{tabular}{|l|l|l|l|l|l|l|l|l|l|l|}
\hline
\multicolumn{2}{|c|}{} & \multicolumn{3}{l}{\makecell{Training}} & \multicolumn{3}{|l|}{\makecell{TestA}} & \multicolumn{3}{l|}{\makecell{TestB}}\\
\cline{3-11}
 \multicolumn{2}{|c|}{}  &\multicolumn{1}{c}{P} & \multicolumn{1}{c}{R} &\multicolumn{1}{l|}{\makecell{F1}}&\multicolumn{1}{c}{P} & \multicolumn{1}{c}{R} &\multicolumn{1}{l|}{\makecell{F1}}&\multicolumn{1}{c}{P} & \multicolumn{1}{c}{R} &\multicolumn{1}{l|}{\makecell{F1}} \\
\hline
\makecell{biLSTM-CRF} &\textit{C} &\multicolumn{1}{c}{\textbf{98.22}} &\multicolumn{1}{c}{96.64}	&\textbf{97.42}&	\multicolumn{1}{c}{69.01}&\multicolumn{1}{c}{63.64}&	66.22&\multicolumn{1}{c}{\textbf{67.35}}&\multicolumn{1}{c}{70.97}	&\textbf{69.11}	\\

&\textit{MW}&\multicolumn{1}{c}{\textbf{91.54}}&\multicolumn{1}{c}{86.44}	&\textbf{88.92}&\multicolumn{1}{c}{\textbf{43.48}}	&\multicolumn{1}{c}{20.83}&	28.17&\multicolumn{1}{c}{\textbf{45.83}}	&\multicolumn{1}{c}{30.14}&	36.36\\
\hline
\multirow{2}{*}{\makecell{biLSTM-CRF + \\ Sch. Matcher}}&\textit{C} &\multicolumn{1}{c}{95.92}&	\multicolumn{1}{c}{\textbf{97.81}}&96.86&\multicolumn{1}{c}{\textbf{74.19}}&	\multicolumn{1}{c}{\textbf{71.88}}&\textbf{73.02}	&\multicolumn{1}{c}{63.29}	&\multicolumn{1}{c}{\textbf{74.63}}	&68.49\\

&\textit{MW}&\multicolumn{1}{c}{87.33}&\multicolumn{1}{c}{\textbf{87.03}}&87.18	&\multicolumn{1}{c}{38.46}&\multicolumn{1}{c}{\textbf{22.73}}&\textbf{28.57}&\multicolumn{1}{c}{42.55}	&\multicolumn{1}{c}{\textbf{32.26}}	&\textbf{36.70}\\
\hline
\end{tabular}
\end{table}

Finally, we test the impact of using the schedule matching together with a biLSTM-CRF network. In this experiment, we consider the network trained using all the features proposed, and the embeddings not pre-trained. Table 8 reports the results obtained. We can observe how generally the system benefits from the use of the schedule information. Especially in the testing part, where the neural network recognizes with less accuracy, the explicit information contained in the schedule can be exploited for identifying the entities at which users are referring while listening to the radio and posting the tweets. 

\section{Conclusion}
We have presented in this work a novel method for detecting musical entities from user-generated content, modelling linguistic features with statistical models and extracting contextual information from a radio schedule. We analyzed tweets related to a classical music radio station, integrating its schedule to connect users' messages to tracks broadcasted. We focus on the recognition of two kinds of entities related to the music field, \textit{Contributor} and \textit{Musical Work}.

According to the results obtained, we have seen a pronounced difference between the system performances when dealing with the \textit{Contributor} instead of the \textit{Musical Work} entities. Indeed, the former type of entity has been shown to be more easily detected in comparison to the latter, and we identify several reasons behind this fact. Firstly, \textit{Contributor} entities are less prone to be shorten or modified, while due to their longness, \textit{Musical Work} entities often represent only a part of the complete title of a musical piece. Furthermore, \textit{Musical Work} titles are typically composed by more tokens, including common words which can be easily misclassified. The low performances obtained in the case of \textit{Musical Work} entities can be a consequences of these observations. On the other hand, when referring to a \textit{Contributor} users often use only the surname, but in most of the cases it is enough for the system to recognizing the entities.

From the experiments we have seen that generally the biLSTM-CRF architecture outperforms the SVM model. The benefit of using the whole set of features is evident in the training part, but while testing the inclusion of the features not always leads to better results. In addition, some of the features designed in our experiments are tailored to the case of classical music, hence they might not be representative if applied to other fields. We do not exclude that our method can be adapted for detecting other kinds of entity, but it might be needed to redefine the features according to the case considered. Similarly, it has not been found a particular advantage of using the pre-trained embeddings instead of the one trained with our corpora. Furthermore, we verified the statistical significance of our experiment by using Wilcoxon Rank-Sum Test, obtaining that there have been not significant difference between the various model considered while testing.  

The information extracted from the schedule also present several limitations. In fact, the hypothesis that a tweet is referring to a track broadcasted is not always verified. Even if it is common that radios listeners do comments about tracks played, or give suggestion to the radio host about what they would like to listen, it is also true that they might refer to a \textit{Contributor} or \textit{Musical Work} unrelated to the radio schedule. 

\subsubsection{Acknowledgments} This work is partially supported by the European Commission under the TROMPA project (H2020 770376), and by the Spanish Ministry of Economy and Competitiveness under the Maria de Maeztu Units of Excellence
Programme (MDM-2015-0502)

%
%
%

\begin{thebibliography}{8}

\bibitem{ref_proc1}
Habib, M.B., Keulen, M.V: Information Extraction for Social Media. SWAIE@COLING (2014) 

\bibitem{ref_proc2}
Ritter, A., Clark, S., Etzioni, O.: Named entity recognition in tweets: an experimental study. In: Proceedings of the 2011 Conference on Empirical Methods in Natural Language Processing, pp. 1524--1534 (2011)

\bibitem{ref_article1}
Nadeau, D., Sekine, S.: A survey of named entity recognition and classification. Lingvisticae Investigationes, \textbf{30}(1), 3--26 (2007)

\bibitem{ref_proc3}
Lample, G., Ballesteros, M., Subramanian, S., Kawakami, K., Dyer, C.: Neural Architectures for Named Entity Recognition. In: Proceedings of NAACL-HLT 2016, pp. 260--270 (2016)

\bibitem{ref_proc4}
Lin, B. Y., Xu, F. F., Luo, Z., Zhu, K. Q.: Multi-channel BiLSTM-CRF Model for Emerging Named Entity Recognition in Social Media. In: Proceedings of the 3rd Workshop on Noisy User-Generated Text, pp. 160--165 (2017)

\bibitem{ref_article2}
Derczynski, L., Maynard, D., Rizzo, G., Van Erp, M., Gorrell, G., Troncy, R., Bontcheva, K.: Analysis of named entity recognition and linking for tweets. Information Processing and Management, \textbf{51}(2), pp. 32--49 (2015)

\bibitem{ref_book1}
M\"uller, M.: Fundamentals of Music Processing. Springer (2015)

\bibitem{ref_article3}
Oramas, S., Ostuni, V. C., Di Noia, T., Serra, X., Di Sciascio, E.: Sound and Music Recommendation with Knowledge Graphs. ACM Transactions on Intelligent Systems and Technology, 8(2), pp. 1--21 (2015)

\bibitem{ref_proc5}
Tata, S., Di Eugenio, B.: Generating Fine-Grained Reviews of Songs from Album Reviews. In: Proceedings of the 48th Annual Meeting of the Association for Computational Linguistics, pp. 1376--1385 (2010)

\bibitem{ref_proc6}
Schedl, M., Hauger, D.: Mining microblogs to infer music artist similarity and cultural listening patterns. In: Proceedings of the 21st International Conference on World Wide Web, pp. 877--886 (2012)

\bibitem{ref_proc7}
Oramas, S., Espinosa-anke, L., Lawlor, A., Serra, X., Saggion, H.: Exploring Customer Reviews for Music Genre Classification and Evolutionary Studies. In: Proceedings of the 17th International Society for Music Information Retrieval Conference, pp. 150--156 (2016)

\bibitem{ref_proc8}
Zhang, X., Liu, Z., Qiu, H., Fu, Y.: A hybrid approach for chinese named entity recognition in music domain. In: Proceedings of the 8th IEEE International Symposium on Dependable, Autonomic and Secure Computing, pp. 677--681 (2009)

\bibitem{ref_proc9}
Oramas, S., Espinosa-Anke, L., Sordo, M., Saggion, H., Serra, X.: ELMD: An Automatically Generated Entity Linking Gold Standard Dataset in the Music Domain. In: Proceedings of the Language Resources and Evaluation Conference, pp. 3312--3317 (2016)

\bibitem{ref_proc10}
 Mendes, P. N., Jakob, M., García-silva, A., Bizer, C.: DBpedia Spotlight : Shedding Light on the Web of Documents. In: Proceedings of the 7th International Conference on Semantic Systems, pp. 1--8 (2011) 

\bibitem{ref_article4}
Ferragina, P., Scaiella, U.: Fast and Accurate Annotation of Short Texts with Wikipedia Pages. IEEE Software, \textbf{29}(1), pp. 70--75 (2012)

\bibitem{ref_article5}
Moro, A., Raganato, A., Navigli, R.: Entity Linking meets Word Sense Disambiguation: a Unified Approach. Transactions of the Association for Computational Linguistics, \textbf{2}(0), pp. 231--244 (2014)

\bibitem{ref_proc11}
Oramas, S., Ferraro, A., Correya, A., Serra, X.:  Mel: a Music Entity Linking System. In: Proceedings of the 18th International Society for Music Information Retrieval Conference (2017)

\bibitem{ref_proc12}
 Zangerle, E., Gassler, W., Specht, G.: Exploiting Twitter's Collective Knowledge for Music Recommendation. In: Proceedings of the WWW'12 Workshop on 'Making Sense of Microposts', pp. 14--17  (2012)

\bibitem{ref_proc13}
Hauger, D., Schedl, M., Košir, A., Tkalcic, M.: The Million Musical Tweets Dataset: What Can We Learn From Microblogs. In: Proceedings of the 14th International Society for Music Information Retrieval Conference (2013)

\bibitem{ref_proc14}
 Zangerle, E., Pichl, M., Gassler, W., Specht, G.: \#nowplaying Music Dataset: Extracting Listening Behavior from Twitter. In: Proceedings of the 1st International Workshop on Internet-Scale Multimedia Management, pp. 21--26 (2014)
 
\bibitem{ref_proc15}
Pennington, J., Socher, R., Manning, C.: Glove: Global Vectors for Word Representation. In: Proceedings of the 2014 Conference on Empirical Methods in Natural Language Processing, pp. 1532--1543 (2014)

\bibitem{ref_article6}
Platt, J.: Fast Training of Support Vector Machines using Sequential Minimal Optimization. Advances in Kernel Methods - Support Vector Learning (1998)

\bibitem{ref_article7}
Frank, E., Hall, M. A., Witten, I. H.: The WEKA Workbench. Online Appendix for "Data Mining: Practical Machine Learning Tools and Techniques", Morgan Kaufmann, Fourth Edition (2016)

\bibitem{ref_article8}
Porcaro, L.,: Information Extraction from User-generated Content in the Classical Music Domain. Master thesis, Pompeu Fabra University, Barcelona, Spain (2018)


\bibitem{ref_proc16}
Reimers, N., Gurevych, I.: Reporting Score Distributions Makes a Difference: Performance Study of LSTM-networks for Sequence Tagging. In: Proceedings of the 2017 Conference on Empirical Methods in Natural Language Processing, pp. 338--348 (2017)


\end{thebibliography}
%

\end{document}